# Belief Updating and Learning in Semi-Qualitative Probabilistic Networks

**Cassio Polpo de Campos**[1,2] **Fabio Gagliardi Cozman**[1]
[1]Escola Politécnica, Univ. de São Paulo, São Paulo, SP - Brazil
[2]Pontifícia Universidade Católica, São Paulo, SP - Brazil
cassio@ime.usp.br, fgcozman@usp.br

## Abstract

This paper explores semi-qualitative probabilistic networks (SQPNs) that combine numeric and qualitative information. We first show that exact inferences with SQPNs are $\mathrm{NP}^{\mathrm{PP}}$-Complete. We then show that existing qualitative relations in SQPNs (plus probabilistic logic and imprecise assessments) can be dealt effectively through multilinear programming. We then discuss learning: we consider a maximum likelihood method that generates point estimates given a SQPN and empirical data, and we describe a Bayesian-minded method that employs the Imprecise Dirichlet Model to generate set-valued estimates.

## 1 Introduction

Qualitative probabilistic networks abstract the precise probability values that are mandatory in Bayesian networks. Instead of displaying precise values, a qualitative probabilistic network (QPN) only states algebraic relations amongst probability values [18, 41]. There are several efficient algorithms for QPNs [18], including algorithms for multiple observations [34], ambiguous signs [4], non-monotonic influences [32] and other relations [5, 6, 31, 35, 38]. Sections 2 reviews the basics of QPNs.

Parsons [28] and Renooij and van der Gaag [33] have proposed semi-qualitative probabilistic networks (SQPN) that mix quantitative and qualitative assessments. For SQPNs the computation of *exact* inferences is generally a more complex undertaking, and existing algorithm focus on approximate solutions [33]. In Section 3 we characterize the complexity of exact inferences for SQPNs, proving that they are $\mathrm{NP}^{\mathrm{PP}}$-Complete.

We then explore, in Section 4, multilinear programming methods that can generate exact inferences in SQPNs, dealing with various influences and synergies without approximations (our methods can also be stopped early to produce approximations). In Section 5 we discuss how probabilistic logic [22] and imprecise [39] assessments can be mixed with numeric and qualitative assessments in the same multilinear programming framework.

Finally, in Section 6 we ask how to use qualitative information when learning from data. We assume that qualitative relations represent *a priori* knowledge, to be combined with a database containing training data. We consider two approaches. The first approach is a maximum likelihood method that views qualitative relations as hard constraints on estimates. The second approach interprets qualitative relations as constraints on prior distributions; we employ the Imprecise Dirichlet Model to capture prior beliefs. These methods deal with an issue that has not received due attention in the literature: namely, how to interpret qualitative information when it must be combined with empirical (quantitative) data.

## 2 Qualitative probabilistic networks

A QPN consists of an acyclic digraph, a set of random variables where each variable is associated with a node in the graph, and a collection of constraints on probability values. The digraph conveys a Markov condition: every variable is independent of its nondescendants nonparents given its parents. We start by assuming that variables are Boolean, with a "higher" value and a "lower" value.

Constraints in a QPN derive from qualitative *influences* and *synergies* amongst probability values [41]. An *influence* between two nodes expresses how the values of one node influence the probabilities of the values of the other node. A *positive influence* of node $A$ on its effect $B$, denoted $S^+(A, B)$, expresses that observing higher values for $A$ makes higher values for $B$ more likely, regardless of any other direct influence on $B$, that is,

$$P(B = b|A = a, x) \geq P(B = b|A = \overline{a}, x) \qquad (1)$$

for any complete instantiation $x$ of $\mathrm{pa}(B) \setminus \{A\}$ — the parents of $B$ are denoted by $\mathrm{pa}(B)$). A *negative influ-* *ence*, denoted by $S^-(A,B)$, and a *zero influence*, denoted by $S^0(A,B)$, are defined analogously, replacing $\geq$ by $\leq$ and $=$ respectively. If the influence of $A$ on $B$ is not monotonic, it is said to be ambiguous, and denoted by $S^?(A,B)$. To simplify notation, we denote an event $\{A = a\}$ by $a$ whenever possible.

*Synergies* represent interactions among influences. An *additive synergy* between three nodes expresses how the values of two nodes jointly influence the probabilities of the values of the third node. A *positive additive synergy* of nodes $A$ and $B$ on their common effect $C$, denoted $Y^+(\{A,B\},C)$, expresses that the joint influence of $A$ and $B$ on $C$ is greater than the sum of their separate influences, regardless of other influences on $C$, that is,

$$P(c|a,b,x) + P\big(c|\overline{a},\overline{b},x\big) \geq P(c|\overline{a},b,x) + P\big(c|a,\overline{b},x\big) \tag{2}$$

for any complete instantiation $x$ of $\mathrm{pa}(C) \setminus \{A,B\}$. *Negative*, *zero*, and *ambiguous additive synergies* are defined analogously. A *product synergy* expresses how the value of a node influences the probabilities of the values of another node given the value of a third node. A *positive product synergy* of node $A$ and node $B$ given $c$, denoted $X^+(\{A,B\},c)$, means:

$$P(c|a,b,x)\,P\big(c|\overline{a},\overline{b},x\big) \geq P\big(c|a,\overline{b},x\big)\,P(c|\overline{a},b,x) \tag{3}$$

for any complete instantiation $x$ of $\mathrm{pa}(C) \setminus \{A,B\}$. *Negative*, *zero*, and *ambiguous product synergies* are defined analogously.

Generally an *inference* in a QPN refers to the qualitative question of how the observation of some variable changes the probability of other variables. Suppose $Q$ is the query variable and $e$ is our observation (an observed event); we need to evaluate $P(q|e) - P(q)$. When $\max P(q|e) - P(q) \leq 0$, we have a negative influence of $e$ in $Q$. If $\min P(q|e) - P(q) \geq 0$, then a positive influence obtains. If both $\max$ and $\min$ are zero, than we have no influence at all. Otherwise, we have an ambiguous influence. A polynomial inference algorithm for QPNs has been proposed by Druzdzel and Henrion [17]: the algorithm generates a sign for each node, implied by the observations in the QPN.

There are several extensions to the QPNs just defined [4, 31, 34]. If the difference $P(b|ax) - P(b|\overline{a}x)$ yields contradictory signs depending on the instantiation $x$ of $\mathrm{pa}(B) \setminus \{A\}$, we say that the influence of $A$ on $B$ is *non-monotonic*. But if we look at this difference separately for each $x$, the influence of $A$ on $B$ is unambiguous, that is, either positive, negative or zero. To capture the sign of a non-monotonic influence for each $x$, Bolt et al. [6] introduced the concept of situational signs. The *positive situational influence* of $A$ on $B$ given some evidence $e$ over $\mathrm{pa}(B) \setminus \{A\}$ is denoted by $S_e^{?(+)}(A,B)$ and indicates that $P(b|ae) \geq P(b|\overline{a}e)$. Negative and zero situational influences are defined analogously. So, for each instantiation $x$ of $\mathrm{pa}(B) \setminus \{A\}$, we may have a different situational influence. Situational signs are a very general way to represent qualitative influences, clearly separating ambiguous influences from non-monotonic ones. Furthermore, Renooij and van der Gaag [31] introduce the "enhanced" formalism for qualitative networks. On this formalism, a *weakly positive influence* of $A$ on its child $B$ indicates that $0 \leq P(b|a,x) - P(b|\overline{a},x) \leq \delta$ for any complete instantiation $x$ of $\mathrm{pa}(B) \setminus \{A\}$, where $\delta$ is a cut-off value. A *strongly positive influence* of $A$ on its child $B$ states that $P(b|a,x) - P(b|\overline{a},x) \geq \delta \geq 0$ for any complete instantiation $x$ of $\mathrm{pa}(B) \setminus \{A\}$. *Weakly negative influence* and *strongly negative influence* are defined analogously.

We adopt a definition of QPN that merges the *situational qualitative networks* of Bolt et al. [6] with the *enhanced qualitative networks* of Renooij and van der Gaag [31]:

**Definition** A qualitative probabilistic network (QPN) consists of an acyclic digraph associated with random variables and a Markov condition, where standard qualitative influences, additive synergies, product synergies, situational influences, weak and strong influences are specified between nodes and parents.

# 3 Semi-qualitative probabilistic networks

A SQPN consists of an acyclic digraph associated with variables and a Markov condition, where each node $A$ is either associated with conditional distributions $P(A|\mathrm{pa}(A))$, or associated with qualitative statements from QPNs (this is a slightly more general definition of SQPN than others found in literature [28, 33]). Thus SQPNs offer a combination of QPNs and Bayesian networks. One might hope that such a combination would not be harder than the hardest of its components; that is, no harder than Bayesian networks. This section shows that SQPNs are harder than QPNs and Bayesian networks, and these two types of networks should be viewed as lower complexity special cases of the former. We start by defining our problem precisely:

**Definition** The decision version of SQPN inference, `D-SQPN-M`, is the problem of finding whether there are probability assignments to all configurations of nodes and their parents that makes $\max P(q|e) - P(q) > 0$.

**Definition** The decision version of SQPN inference, `D-SQPN-m`, is the problem of finding whether there are probability assignments to all configurations of nodes and their parents that makes $\min P(q|e) - P(q) < 0$.

In both cases $Q$ is the query node and $\{E = e\}$ is our observation in the network. Clearly to solve the qualitative inference we need (and it is enough) to answer both `D-SQPN-M` and `D-SQPN-m` problems.

**Theorem 1** `D-SQPN-M` *and* `D-SQPN-m` *are* $NP^{PP}$*-Complete.*

**Proof** First, note that D-SQPN-m belongs to $NP^{PP}$ because given probability assignments to all configurations of nodes and their parents, we obtain a standard Bayesian network. In this case, the calculation of $\min P(q|e) - P(q)$ can be made by the PP oracle.

To show hardness of D-SQPN-m, we reduce the EMAJSAT problem (following previous work by Park and Darwiche [27]): *Given a Boolean formula $\phi$ over a set of variables $\{X_1, \ldots, X_n\}$, and an integer $1 \leq k \leq n$, is there an assignment for the variables $X_1, \ldots, X_k$ under which the majority of worlds satisfy $\phi$?* Let $\mathbb{X}$ represent the first $k$ variables, that is, $X_1, \ldots, X_k$ and $\mathbb{Y}$ the others, that is, $X_{k+1}, \ldots, X_n$. We construct a SQPN modeling the formula $\phi$. This network has a qualitative node for each variable in $\mathbb{X}$ with no parent; the $\mathbb{Y}$ variables have no parents and uniform prior probability. Furthermore, there is a node $W_i$ for each Boolean operator. The parents of any operator $W_i$ are its operands in the formula and $P(w_i|\mathrm{pa}(W_i))$ encodes its truth table. Let $W_0$ be the only operator without children in the network. Insert a dummy binary child to it, named $Q$, with $W_0$ and a new qualitative node $E$ as parents (see Figure 1). We impose $P(q|w_0, e) = \frac{1}{2}$, $P(q|w_0, \overline{e}) = 1$, $P(q|\overline{w_0}, e) = \frac{1}{2}$ and $P(q|\overline{w_0}, \overline{e}) = 0$. An inference in the SQPN on the influence of $\{E = e\}$ over $Q$ will solve EMAJSAT. To answer this question, we need to evaluate the sign of $\min P(q|e) - P(q)$. Note that $P(q|e) = P(q|w_0, e)\, P(w_0) + P(q|\overline{w_0}, e)\, P(\overline{w_0}) = \frac{1}{2}$ and $P(q) = P(w_0)\,(1 - P(e)) + \frac{P(e)}{2}$. Thus $P(w_0) > \frac{1}{2}$ implies $\min P(q|e) - P(q) < 0$ and $P(w_0) \leq \frac{1}{2}$ implies $\min P(q|e) - P(q) = 0$ (because $P(e)$ will be set to 1 in this case). Suppose $P(w_0)$ counts the number of $\mathbb{Y}$ worlds that satisfy $\phi$; then it is only necessary to compute the answer of the qualitative query: if $\min P(q|e) - P(q) < 0$, then we have found an instantiation for $\mathbb{X}$ variables that satisfy the EMAJSAT requirements, that is, where the majority of worlds of $\mathbb{Y}$ satisfy $\phi$. Otherwise there is no such instantiation for the $\mathbb{X}$ variables. It remains to show that $P(w_0)$ counts the number of $\mathbb{Y}$ possible worlds satisfying $\phi$ given the instantiation for $\mathbb{X}$ (then the query $\min \frac{1}{2} - P(q)$ will maximize $P(w_0)$, finding such instantiation for the $\mathbb{X}$ variables). We have

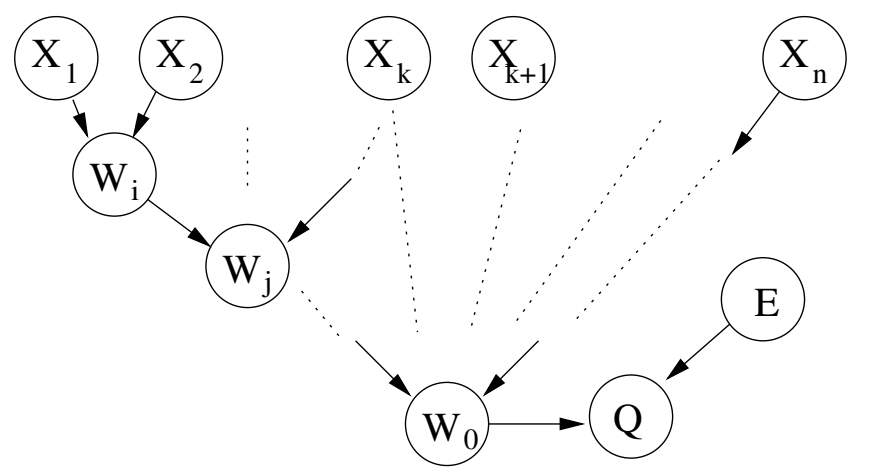

Figure 1: Network used in Theorem 1.

$P(w_0) = \sum_{X,Y} P(w_0|X, Y)\, P(X)\, P(Y)$ and then

$$P(w_0) = \sum_{Y} \left( P(w_0|xY)\, \frac{1}{2^{n-k}} \right) = \frac{|\mathrm{sat}|}{|\mathrm{total\ cases}|}.$$

Note that the previous summation over $X$ disappeared because there is only a term where $P(X)$ equals 1 (it equals 0 otherwise). We indicate that instantiation by $\{X = x\}$. Furthermore, $P(w_0|xY)$ is 1 when $(x, Y)$ satisfy $\phi$, and 0 otherwise. This implies that $\max P(w_0)$ evaluates which instantiation of $\mathbb{X}$ has the greater fraction of $\mathbb{Y}$ worlds satisfying $\phi$. The proof for D-SQPN-M is analogous. □

We have used in Theorem 1 a network with very simple qualitative relations to obtain $NP^{PP}$-Hardness; the inclusion of other qualitative influences and synergies, situational signs and non-monotonic relations, can only make the problem harder, but the problem still belongs to $NP^{PP}$. This implies that exact inferences in more specialized semi-qualitative networks [28, 33] are $NP^{PP}$-Complete too.

## 4 Inferences in SQPNs through multilinear programming

A semi-qualitative inference can be formulated as a non-linear programming problem. The goal is to minimize/maximize the expression $P(q|e) - P(q)$, subject to numeric assessments and qualitative relations; depending on the signs produced in minimization/maximization, we know how $e$ influences $Q$. The query can be written using a (multilinear) expression from Bayesian network theory:

$$P(q|e) = \sum_{Y \in \{X_i\}} \prod_{Y \in \{X_i, Q\}} p(Y|\mathrm{pa}(Y), e)\,, \qquad (4)$$

where $X_i \notin \{Q, E\}$. As described in Sections 2 and 3, all qualitative restrictions can be written as linear and multilinear constraints.

The difficulty is that Expression (4) potentially contains a huge number of multilinear terms (the number of terms is exponential on the size of the network). We propose to transform Expression (4) into a collection of smaller constraints. Our solution method is heavily based on a previous algorithm we have developed for credal networks [15], so we describe rather briefly the version for SQPNs here.

Imagine that a variable elimination algorithm were run in a Bayesian network with the same structure of the SQPN of interest. The idea is to run variable elimination and, as a bucket tree is generated, to construct an optimization problem with multilinear and linear constraints for each bucket [10]. That is, the algorithm follows each step of a variable elimination algorithm but it stores calculations as symbolic multilinear constraints. Each one of these multilinear and linear constraints represents local information in the network; that is, constraints represent relations between neighbour buckets in the tree. The number of functions in this new multilinear programming problem depends mainly on the topology of the network. The transformation procedure can be quickly executed as its complexity is on the order of a Bayesian network inference.

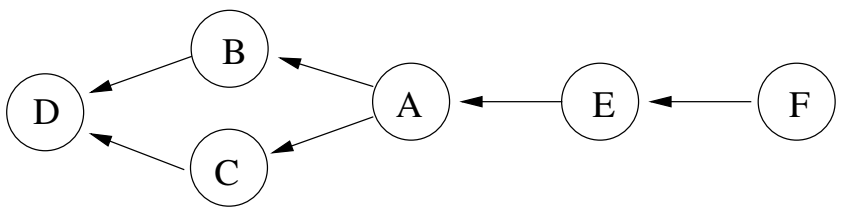

Figure 2: A simple example network.

**Example 1** To illustrate the transformation, take a simple network with topology shown in Figure 2. Any variable $X \in \{A, B, C, D, E, F\}$ assumes values $\{x_0, x_1\}$. Suppose we need to evaluate the influence of $f_0$ on $D$; that is, we must compute the sign of $p(d_0|f_0) - p(d_0)$. This could be computed by solving a "long" version of the problem:

$$\begin{aligned} \min / \max \sum_{h,i,j,k} & p(d_0|b_i c_j)\, p(b_i|a_k)\, p(c_j|a_k) \times \\ & \times p(a_k|e_h)\, p(e_h|f_0) - \sum_{g,h,i,j,k} p(d_0|b_i c_j)\, p(b_i|a_k) \times \\ & \times p(c_j|a_k)\, p(a_k|e_h)\, p(e_h|f_g)\, p(f_g) \qquad (5) \end{aligned}$$

subject to quantitative and qualitative constraints. We have a multilinear objective function with 48 nonlinear terms of degree five and six. Instead of dealing with this function, we can transform it into a problem with simpler multilinear functions of degree at most two, by grouping terms and introducing new variables. Note that $p(d_0|f_0) - p(d_0)$ equals $p(d_0|f_0)\, p(f_1) - p(d_0|f_1)\, p(f_1)$ and as long as we are interested in its sign, we can divide it by $p(f_1)$ without change in the resulting sign. We get a multilinear program with just 36 nonlinear terms in place of those 48 needed before: $\min / \max \; p(d_0|f_0) - p(d_0|f_1)$ subject to

$p(d_0|f_k) = \sum_j p(d_0|b_j, f_k)\, p(b_j|f_k)$ , for $k = 0, 1$,
$p(d_0|b_i, f_k) = \sum_j p(d_0|b_i, c_j)\, p(c_j|f_k)$ , for $i, k = 0, 1$,
$p(b_i|f_k) = \sum_j p(b_i|a_j)\, p(a_j|f_k)$ , for $i, k = 0, 1$,
$p(c_i|f_k) = \sum_j p(c_i|a_j)\, p(a_j|f_k)$ , for $i, k = 0, 1$,
$p(a_i|f_k) = \sum_j p(a_i|e_j)\, p(e_j|f_k)$ , for $i, k = 0, 1$,

plus the quantitative and qualitative constraints. This reduction in the number of multilinear terms is exponentially more effective as the size of the network grows.

Other authors [1, 19] have also proposed transformations that produce multilinear programs for graphical models. These previous proposals have worked with the space of *possible worlds*, whose size is exponential on the parameters of the network. The advantage here is the direct formulation, in the sense that generation of the multilinear problem follows the same method used in Bayesian network inference. Due to this, it is possible to develop specialized approximate algorithms for the problem, obtaining tight variable bounds which improve the multilinear solver performance (more details in [13, 15]).

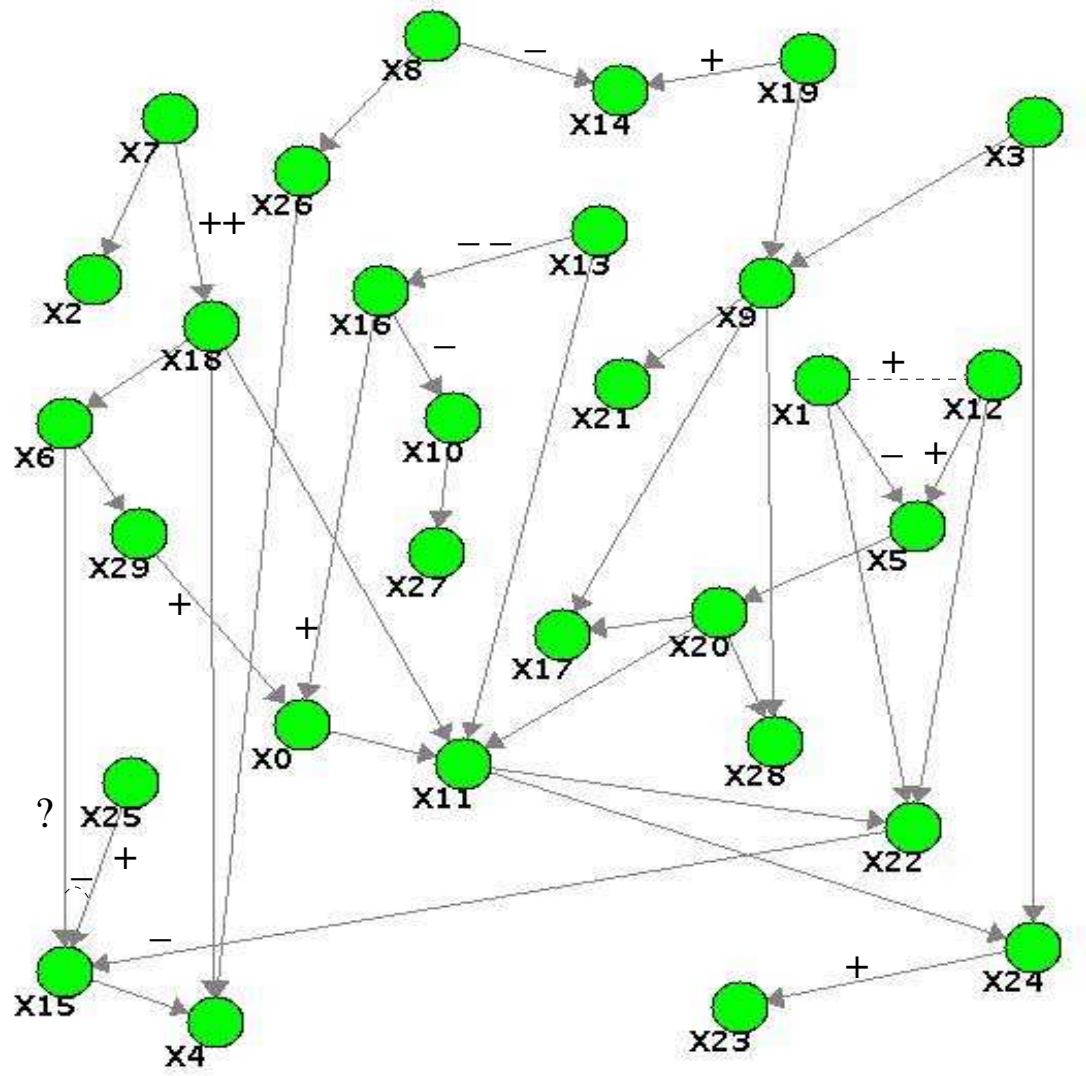

Figure 3: A semi-qualitative network.

To solve the multilinear problem, we have used Sherali and Tuncbilek's Reformulation-Linearization (RL) method [36]. The RL method substitutes each product of variables $\prod_{j \in J_{rt}} \theta_j$ by a new artificial variable $\vartheta_{J_{rt}}$ for all terms $t$ in the problem, thus obtaining a linear program. The solution of each linear problem gives an upper bound to the solution of the multilinear problem. The method iterates over the variables by branching over their ranges whenever necessary, until each $\vartheta_{J_{rt}}$ is close enough to $\prod_{j \in J_{rt}} \theta_j$.

We would like to emphasize that this method is quite effective in practice, and not just a theoretical exercise. We have been able to easily handle inferences in networks containing up to 150 nodes and a substantial amount of qualitative relations (using random test networks [24]).

**Example 2** To illustrate the multilinear programming method, consider the network in Figure 3. Nodes without qualitative signs are associated with probability distributions. The multilinear scheme (implemented in the C language with calls to CPLEX solver) can process this network almost instantly in a Pentium IV computer. For instance, the algorithm finds that observation $\{X_{13} = x_{13}\}$ influences $X_4$ negatively.

# 5 Propositional probabilistic logic, imprecise assessments, and credal networks

Given that Boolean variables are generally at our disposal in SQPNs,[1] we can easily mix logical constraints with probabilistic assessments. Actually, we can take a further step, exploring assessments that simultaneously in-clude logical and probabilistic elements. For example, we may have the conditional assessment $P(A \vee \neg B|C) = \alpha$. Such assessments belong to propositional *probabilistic logic* [26], and they have rarely been combined with statistical graphical models — inference algorithms in propositional probabilistic logic usually reduce all assessments to linear programming by carefully avoiding the nonlinear character of independence relations [37, 21, 22].

[1]We have assumed that variables are Boolean, as qualitative relations are usually discussed in this case. Results discussed so far directly apply to non-Boolean settings.

In our multilinear programming method, assessments such as $P(A \vee \neg B|C) = \alpha$ (a linear constraint) can be freely combined with independence relations to produce multilinear constraints. Clearly the complexity of inferences depends on the number of variables involved in assessments; if we limit the assessments to variables and their parents, we stay within the same complexity level as SQPNs.

**Example 3** Consider the network in Example 2. The constraint $P(X_{18} \vee \neg X_7|X_2 = x_2) = 0.95$ does affect the inference, causing the influence of $\{X_{13} = x_{13}\}$ on $X_4$ to change from -0.092 to -0.083.

Imprecise assessments can also be directly folded into a multilinear framework; that is, a numeric assessment may state that a probability distribution belongs to a set of distributions [39] or to some "order-of-magnitude" representation [14, 30]. Sets of probability distributions are called *credal sets* [25], and statistical graphical models associated with credal sets have been called *credal networks* [12, 20]. A credal network consists of an acyclic digraph where each node is associated with a variable, and each variable $X_i$ is associated with a collection of credal sets $K(X_i|\text{pa}(X_i))$. As SQPNs form a subclass of credal networks, the proof of Theorem 1 can be used to show that inference in credal networks is $\text{NP}^{\text{PP}}$-Hard.[2] As pertinence in $\text{NP}^{\text{PP}}$ is immediate for inferences with credal networks, we obtain a proof for $\text{NP}^{\text{PP}}$-Completeness of marginal inferences in credal networks (this result was already sketched in [11]). The remarkable point here is that *exact* inference in SQPNs can lead us, in a worst-case sense, to face the same complexity of exact inference in credal networks.

# 6 Learning with semi-qualitative a priori information and empirical data

Suppose an expert represents her beliefs about a set of variables through a SQPN. Later, the variables in the SQPN are observed several times. How to combine the beliefs of the expert with the empirical data? There is relatively little guidance in the literature on how to do such combination. We first discuss an intuitive maximum likelihood method based on constraints on estimates; after identifying some drawbacks of this method, we move to Bayesian-minded method based on constraints on priors.

[2]The only difference between the problem constructed in Theorem 1 and a marginal inference in a credal network is the objective function; if we compute $\max P(w_0)$ directly (this is already a marginal inference in a credal network) instead of $\min P(q|e) - P(q)$, we obtain the desired result for credal networks.

The following notation is used in the remainder of this section. The probability value $P(X_i = x_{ij}|\text{pa}(X_i) = \pi_{ik})$ is denoted by $\theta_{ijk}$, and $N_{ijk}$ is the number of occurrences of configuration $\{X_i = x_{ij}, \text{pa}(X_i) = \pi_{ik}\}$. The set of $\theta_{ijk}$ for a given $i$ is denoted by $\theta_i$; the set of $\theta_{ijk}$ for given $(i, k)$ is denoted by $\theta_{ik}$; the set of all $\theta_{ijk}$ is denoted by $\theta$. We denote by $Q_{im}$ the $m$th qualitative relation assessed for $\theta_i$. Finally, we assume that a database $D$ is available containing $N$ complete records $D_l$ (no missing data).

## 6.1 A maximum likelihood method

The likelihood $L(\theta)$ for data $D$ is $L(\theta) = P(D|\theta) = \prod_l P(D_l|\theta) = \prod_{i,j,k} \theta_{ijk}^{N_{ijk}}$ [9]. Suppose for a moment that only qualitative relations are present in the available SQPN. A reasonable maximum likelihood estimate of $\theta$ would then be $\hat{\theta} = \arg\max_\theta \prod_{i,j,k} \theta_{ijk}^{N_{ijk}}$, subject to $Q_{im}$ for all $(i, m)$ and $\sum_j \theta_{ijk} = 1$ for all $(i, k)$ [42]. This optimization involves a polynomial objective function subject to multilinear constraints. This potentially large program can be decomposed in $n$ smaller programs:

$$\hat{\theta}_i = \arg\max_{\theta_i} \prod_{j,k} \theta_{ijk}^{N_{ijk}}, \tag{6}$$

subject to $Q_{im}$ for all $m$, and $\sum_j \theta_{ijk} = 1$ for all $k$. The size of program (6) is polynomial on the family of $X_i$ (that is, the number of configurations for $\{X_i, \text{pa}(X_i)\}$). Note that program (6) must be constructed only for those nodes that are associated with qualitative relations; nodes that are "free" can be independently processed by usual maximum likelihood expressions — thus their estimates will be the relative frequencies observed in $D$.

**Example 4** Consider a fragment of the network in Example 2, formed by variables $X_5$, $X_1$ and $X_{12}$. To simplify notation, denote $X_5$ by $X$, $X_{12}$ by $Y$ and $X_1$ by $Z$. We are interested in $\theta_{YZ}$, the probability of $\{X = x\}$ conditional on $Y$ and $Z$. Suppose that a database with $N = 40$ realizations of all variables is collected, and the following counts summarize realizations of $X$ given values of $Y$ and $Z$:[3]

| $Y$ | $Z$ | $N_x$ | $N_{\overline{x}}$ |
|---|---|---|---|
| $y$ | $z$ | 3 | 3 |
| $y$ | $\overline{z}$ | 6 | 14 |

| $Y$ | $Z$ | $N_x$ | $N_{\overline{x}}$ |
|---|---|---|---|
| $\overline{y}$ | $z$ | 8 | 2 |
| $\overline{y}$ | $\overline{z}$ | 1 | 3 |

We must maximize the likelihood $L(\theta)$:

$$\theta_{yz}^3(1-\theta_{yz})^3\theta_{y\overline{z}}^6(1-\theta_{y\overline{z}})^{14}\theta_{\overline{y}z}^8(1-\theta_{\overline{y}z})^2\theta_{\overline{y}\overline{z}}(1-\theta_{\overline{y}\overline{z}})^3$$

subject to $\theta_{yz} \leq \theta_{\overline{y}z}$, $\theta_{y\overline{z}} \leq \theta_{\overline{y}\overline{z}}$, $\theta_{yz} \geq \theta_{y\overline{z}}$, $\theta_{\overline{y}z} \geq \theta_{\overline{y}\overline{z}}$, $\theta_{yz}\theta_{\overline{y}\overline{z}} \geq \theta_{y\overline{z}}\theta_{\overline{y}\overline{z}}$, and $\theta_{YZ} \geq 0$, given relations $S^+(Y, X)$,

[3]Data coming from a distribution where $\theta_{yz} = 1/2$, $\theta_{y\overline{z}} = 1/4$, $\theta_{\overline{y}z} = 3/4$, and $\theta_{\overline{y}\overline{z}} = 1/2$.

$S^{-}(Z, X)$ and $X^{-}(\{Y, Z\}, X = x)$. We then used the MINOS package to compute $\hat{\theta}_{yz} = 0.56$, $\hat{\theta}_{y\overline{z}} = 0.78$, $\hat{\theta}_{\overline{y}z} = 0.27$, and $\hat{\theta}_{\overline{yz}} = 0.37$ (estimates are not equal to relative frequencies, as the latter violate qualitative constraints). A suitably modified EM algorithm might also have been used here [42].

### 6.2 SQPNs as constraints on prior distributions

The maximum likelihood method described in the previous section is rather intuitive in its interpretation of qualitative relations. However the method faces some conceptual difficulties, apart from its computational challenges.

First, the maximum likelihood method deals only with qualitative relations in the SQPN — what to make of possible *numeric* assessments in the SQPN? The expert may wish to indicate precise probability values in the network; these probabilities cannot be directly taken as constraints (if they are fixed at this stage, the empirical data cannot change them). Or the expert may announce a prior distribution for some $\theta_{ijk}$; how should this distribution be integrated into the estimates? These questions are difficult to answer because prior distributions have no clear place within maximum likelihood methods.

Second, the interpretation of qualitative relations as "hard" constraints on estimates is perhaps too inflexible. The expert is required to state precise boundaries between "feasible" and "unfeasible" values of $\theta$; how could an expert learn of such sharp boundaries in general? Now, suppose the expert states an incorrect qualitative relation — as more and more data are collected, we obtain more and more evidence that the expert is mistaken (as the likelihood $L(\theta)$ concentrates around the "true" value of $\theta$) and yet there is little we can do short of throwing the expert's opinion away.

It is perhaps more profitable to interpret the expert's SQPN as a *possibly partial assessment of a prior distribution* over $\theta$. Thus any qualitative relation might be viewed as a constraint on prior distributions, and numeric assessments would be readily interpreted as prior means (or some similarly meaningful measure). Empirical data would be processed by combining the likelihood $L(\theta)$ with whatever prior distributions satisfy the constraints in the SQPN. The next section offers our main contribution concerning learning, as a new parametric formulation for these qualitative "Bayesian" priors.[4]

[4] An alternative non-parametric formulation would take the set of all prior distributions that satisfy qualitative relations, and proceed from that set. The drawback of this otherwise attractive idea is that qualitative relations are rather weak constraints; the set of prior distributions satisfying them is usually too large — and vacuous priors can only produce vacuous posteriors, so no learning obtains [39]. A parametric approach seems more appropriate.

### 6.3 The constrained Imprecise Dirichlet Model

We start assuming that an expert has specified a SQPN, denoted by $\mathcal{N}_p$, that conveys her prior beliefs. Our goal is to learn the parameters of multinomial distributions on $\theta_{ik}$ using both $\mathcal{N}_p$ and data. We select Dirichlet distributions as a natural parametric model for $P(\theta_{ik})$, because the Dirichlet distribution is conjugate with the multinomial distribution [16]. A possible parameterization is $P(\theta_{ik}) \propto \prod_j \theta_{ijk}^{st_{ijk}-1}$ for $s \geq 0$ and $\sum_j t_{ijk} = 1$ where the hyperparameter $s$ controls dispersion and hyperparameters $t_{ijk}$ control location [40]. The parameter $s$ is often interpreted as the "size" of a database encoding the same prior beliefs as the distribution [23]. We assume that $\mathcal{N}_p$ is associated with a single positive number $s_p$ that encodes the "quality" of the SQPN (intuitively, $s_p$ is the size of a database conveying as much prior information as $\mathcal{N}_p$).

Our proposal is to view the content of $\mathcal{N}_p$ as *constraints on the hyperparameters* $t_{ijk}$ of Dirichlet distributions for a fixed value of $s$. To emphasize: the qualitative constraints on the SQPN are *constraints on the prior*, not constraints of the probability values themselves. Our proposal can be stated in more detail as follows.

1. Consider the *numeric* assessments in $\mathcal{N}_p$. For each variable $X_i$ such that $\{\mathrm{pa}(X_i) = \pi_k\}$ is a configuration associated with numeric assessments, define a Dirichlet distribution with hyperparameters $s_p$ and $\tau_{ijk}$, where $\tau_{ijk}$ are the assessments in $\mathcal{N}_p$. Thus we adopt a standard Bayesian model for the quantitative part of the SQPN [23].

2. Now consider the *qualitative* relations in $\mathcal{N}_p$. For each variable $X_i$ associated with qualitative relations, define a *set* of distributions containing all Dirichlet distributions with hyperparameter $s_p$ and hyperparameters $t_{ijk}$ satisfying qualitative relations for $X_i$.

In short, we associate a *single* Dirichlet distribution with numeric assessments in $\mathcal{N}_p$, and a *set* of Dirichlet distributions with qualitative relations in $\mathcal{N}_p$. We should note that sets of Dirichlet distributions with fixed hyperparameter $s$ have received great attention recently [3, 29, 43]. The model is usually referred to as the *Imprecise Dirichlet Model* (IDM) and was introduced by Walley as a model for states of ignorance where prior probabilities can vary freely between 0 and 1 [40]. Our contribution here has been to bring the IDM to the setting of qualitative relations, and to interpret these relations as constraints on hyperparameters, a move that has not yet been pursued in the literature.

We assume, to compute estimates $\hat{\theta}$, that joint prior distributions for all values in $\theta$ are obtained by the product of prior distributions for each $\theta_{ik}$ (as usually assumed in Bayesian networks [23]). We also assume that estimates should minimize the sum of quadratic losses for all $\theta_{ijk}$.

For the quantitative parts of the SPQN, these assumptions lead to a purely Bayesian solution: the optimal estimate $\hat{\theta}_{ijk}$ is then the posterior expected value $E[\theta_{ijk}]$, and the estimates are [16]:

$$\hat{\theta}_{ijk} = (s\tau_{ijk} + N_{ijk})/(s + \sum_j N_{ijk}). \qquad (7)$$

The result for those configurations that are associated with the constrained IDM is not a standard Bayesian solution. However, the estimates also come directly from the conjugacy properties of the Dirichlet and multinomial distributions. For each valid set of hyperparameters $t_{ijk}$, the estimates are again given by Expression (7). Thus we obtain a *set* of estimates defined by (7) and subject to whatever multilinear constraints are imposed by the qualitative relations.

**Example 5** Consider the network fragment and data in Example 4. Suppose the expert adds numeric assessments to the SQPN: $P(Y = y) = 2/3$ and $P(Z = z) = 1/4$; suppose also the hyperparameter $s_p$ is fixed at 2 (following proposals in the literature [3]). Finally, suppose the following counts summarize data on $Y$ and $Z$: $N_y = 29$ and $N_z = 25$ (while $N_y$ is close to its expected value, $N_z$ suggests a possible mistake by the expert). Estimates are: $\hat{\theta}_y = 0.72$, $\hat{\theta}_z = 0.61$, $\hat{\theta}_{yz} = (2t_{yz}+3)/8$, $\hat{\theta}_{y\overline{z}} = (2t_{y\overline{z}}+6)/22$, $\hat{\theta}_{\overline{y}z} = (2t_{\overline{y}z}+8)/12$, and $\hat{\theta}_{\overline{yz}} = (2t_{\overline{yz}}+1)/6$, where $t_{yz} \leq t_{\overline{y}z}$, $t_{y\overline{z}} \leq t_{\overline{yz}}$, $t_{yz} \geq t_{y\overline{z}}$, $t_{\overline{y}z} \geq t_{\overline{yz}}$, $t_{yz}t_{\overline{yz}} \geq t_{y\overline{z}}t_{\overline{yz}}$, and $t_{YZ} \geq 0$. Inference requires multilinear programming; for instance, $P(X = x) \in [0.21, 0.32]$ with the estimates just indicated.

This example has left the estimates as a function of "free" parameters $t_{ijk}$. One possibility would be to algebraic eliminate the hyperparameters $t_{ijk}$ — this is trivial because the relationship between estimates and $t_{ijk}$ is linear. However, there is no need to eliminate $t_{ijk}$; they can be carried as a part of the resulting network. The net result is that we have a simple learning method that produces set estimates — thus the resulting structure is a credal network. The same comment applies when interval or set constraints are imposed over $t_{ijk}$.

This formulation has several attractive features. First, it deals with qualitative and numeric aspects of SQPNs in a uniform manner. Second, it uses constraints only on priors, thus mistakes incurred by the expert can be eventually corrected with enough data. Third, a single hyperparameter $s_p$ must be elicited to capture the quality of the prior. Fourth, the computation of estimates can be done in a readily and efficient manner.

As discussed in Section 4, inferences for rather large qualitative/credal networks can be generated by multilinear programming. Densely connected networks may still present computational challenges; in those cases one may resort to approximate inference [7, 8, 13].

In closing, we note that point estimates can also be generated from constrained IDM priors. The $\Gamma$-minimax strategy is to look for point estimates that minimize maximum loss [2]. Due to the computational challenges posed by such a minimax problem, we have chosen not to follow this path.

# 7 Conclusion

We can summarize the contributions of this paper as follows.

First, we have characterized the complexity of *exact* inference in SQPNs. Theorem 1 shows that, as far as exact inference is concerned, the worst-case behavior of SQPNs and credal networks is the same, so we can suspect that inference algorithms in both classes can benefit from each other. As some SQPNs are just as hard as credal networks, we should be prepared to employ multilinear programming methods when facing the "hardest" SQPNs.

Second, we have presented an inference method, based on multilinear programming, that can encompass existing qualitative relations in SQPNs and several other numeric (precise and imprecise) assessments. We would like to suggest that this general multilinear programming framework can be a useful meeting point for several related uncertainty formalisms — it certainly is a viable approach that can be used in practice.

Third, we have explored the combination of SQPNs and empirical data — an issue of clear practical importance and yet little explored at the moment. We have discussed two approaches: a maximum likelihood method and a Bayesian method based on the Imprecise Dirichlet Model (IDM). Despite the conceptual simplicity of the maximum likelihood approach, we feel that the IDM is far superior: its formulation is compact, and inferences are obtained in closed form. However, the result is a credal network, not a Bayesian network — inferences must then involve some form of exact or approximate multilinear programming. The fact that empirical (quantitative) data moves us from a prior SQPN towards more general credal networks again suggests that in practice we may have to deal with multilinear programming methods.

# Acknowledgements

This work has received generous support from HP Brazil R&D. The work has also been supported by CNPq, CAPES and FAPESP.